\documentclass[conference]{IEEEtran}
\IEEEoverridecommandlockouts

\usepackage{cite}
\usepackage{amsmath,amssymb,amsfonts}
\usepackage{algorithmic}
\usepackage[ruled,vlined]{algorithm2e} % Added for pseudo-code
\usepackage{graphicx}
\usepackage{textcomp}
\usepackage{xcolor}
\usepackage{booktabs}
\usepackage{multirow} % Added for complex tables
\usepackage{hyperref}

\usepackage{tikz}
\usepackage{pgfplots}
\pgfplotsset{compat=1.18}
\usetikzlibrary{shapes.geometric, arrows, positioning}

\def\BibTeX{{\rm B\kern-.05em{\sc i\kern-.025em b}\kern-.08em
    T\kern-.1667em\lower.7ex\hbox{E}\kern-.125emX}}

\begin{document}

\title{Chunking, Retrieval, and Re-ranking: An Empirical Evaluation of RAG Architectures for Policy Document Question Answering}

\author{
    \IEEEauthorblockN{Anuj Maharjan\textsuperscript{*}\thanks{\textsuperscript{*}Both authors contributed equally to this work.}}
    \IEEEauthorblockA{\textit{Electrical Engineering and Computer Science (EECS)} \\
    \textit{University of Toledo}\\
    Toledo, OH, USA \\
    anjmhrjn1@gmail.com}
    \and
    \IEEEauthorblockN{Umesh Yadav\textsuperscript{*}}
    \IEEEauthorblockA{\textit{Electrical Engineering and Computer Science (EECS)} \\
    \textit{University of Toledo}\\
    Toledo, OH, USA \\
    yadav.umesh0518@gmail.com}
}

\maketitle

\begin{abstract}
The integration of Large Language Models (LLMs) into the public health policy sector offers a transformative approach to navigating the vast repositories of regulatory guidance maintained by agencies such as the Centers for Disease Control and Prevention (CDC). However, the propensity for LLMs to generate hallucinations, defined as plausible but factually incorrect assertions, presents a critical barrier to the adoption of these technologies in high-stakes environments where information integrity is non-negotiable \cite{zhang2023siren}. This empirical evaluation explores the effectiveness of Retrieval-Augmented Generation (RAG) architectures in mitigating these risks by grounding generative outputs in authoritative document context \cite{gao2023retrieval}. Specifically, this study compares a baseline Vanilla LLM against Basic RAG and Advanced RAG pipelines utilizing cross-encoder re-ranking. The experimental framework employs a Mistral-7B-Instruct-v0.2 model \cite{jiang2023mistral} and an all-MiniLM-L6-v2 embedding model \cite{wang2020minilm} to process a corpus of official CDC policy analytical frameworks and guidance documents. The analysis measures the impact of two distinct chunking strategies, recursive character-based and token-based semantic splitting, on system accuracy, measured through faithfulness and relevance scores across a curated set of complex policy scenarios \cite{es2023ragas}. Quantitative findings indicate that while Basic RAG architectures provide a substantial improvement in faithfulness (0.621) over Vanilla baselines (0.347), the Advanced RAG configuration achieves a superior faithfulness average of 0.797. These results demonstrate that two-stage retrieval mechanisms are essential for achieving the precision required for domain-specific policy question answering, though structural constraints in document segmentation remain a significant bottleneck for multi-step reasoning tasks.
\end{abstract}


\section{Introduction}

The current landscape of natural language processing is dominated by transformer-based Large Language Models that have demonstrated remarkable capabilities in text synthesis, summarization, and zero-shot reasoning \cite{vaswani2017attention, devlin2019bert}. Despite these advancements, the utility of standalone models in specialized domains like public health policy is severely hampered by their reliance on internal weights that may contain outdated, biased, or incorrect information \cite{gao2023retrieval}. In the context of regulatory compliance and policy analysis, a single erroneous recommendation can have profound implications for public safety and institutional credibility. Consequently, there is an urgent need for architectural frameworks that can reliably bridge the gap between generative fluency and factual grounding \cite{guan2025privacy}.

The motivation behind this research stems from the increasing complexity of government policy repositories. Public health officials are frequently required to synthesize information across disparate documents, such as the CDC Policy Analytical Framework, Strategy and Policy Development guidelines, and Program Cost Analysis manuals. Manually navigating these documents is time-intensive and prone to human error, yet traditional keyword-based search engines lack the semantic depth to answer complex, scenario-based queries. Retrieval-Augmented Generation (RAG) has emerged as a promising paradigm that combines the strengths of information retrieval with the generative capacity of LLMs by injecting relevant document fragments directly into the model’s prompt \cite{lewis2020retrieval}.

The fundamental research question addressed in this paper is: To what extent do advanced retrieval techniques, such as cross-encoder re-ranking and optimized chunking strategies, improve the faithfulness and relevance of generated answers in a policy-specific question-answering system? By evaluating these architectures using a specialized corpus of CDC policy documents, this study seeks to provide a reproducible baseline for the deployment of intelligent policy navigators in regulated information environments. The significance of this work lies in its empirical comparison of system configurations, highlighting the critical role of the retrieval pipeline in ensuring that LLMs act not as repositories of knowledge, but as precise synthesizers of authoritative evidence.

\section{State of the Field and Comparative Literature}

The concept of Retrieval-Augmented Generation was first formalized by Lewis et al. (2020) \cite{lewis2020retrieval} as a method to provide LLMs with access to a non-parametric memory in the form of a dense vector index. This architecture represented a significant departure from previous approaches that relied solely on supervised fine-tuning to encode domain-specific knowledge into model parameters.

\subsection{The Evolution of Retrieval Mechanisms}

The retrieval stage of RAG has traditionally relied on bi-encoder architectures, which independently encode queries and document chunks into a joint vector space. Models like Sentence-BERT (SBERT) \cite{reimers2019sentence} revolutionized this space by enabling efficient semantic similarity searches using cosine similarity, reducing the time required to find relevant passages from hours to milliseconds. However, subsequent evaluations have identified significant limitations in bi-encoders, particularly their inability to capture token-level interactions between a query and a document. This "shallow" retrieval often results in the selection of passages that are semantically similar in a broad sense but contextually irrelevant to the specific intent of a query \cite{liu2023lost}.

To address this, researchers have introduced two-stage retrieval systems that utilize cross-encoders for re-ranking. Unlike bi-encoders, cross-encoders jointly process the query and each candidate document, allowing the transformer’s attention mechanism to evaluate the precise relationship between tokens. Benchmarks on the MS MARCO dataset have shown that re-ranking can improve retrieval precision by as much as 27\% in terms of Mean Reciprocal Rank (MRR@10) \cite{nogueira2019passage}.

\subsection{Legal and Regulatory NLP}
Beyond general retrieval, specific challenges exist in the legal and regulatory domain. Evaluating LLMs on legal benchmarks (LegalBench) has shown that general-purpose models often fail to distinguish between "mandatory" and "permissive" language (e.g., "must" vs "should") \cite{pipitone2024legalbench}. Domain-specific RAG adaptation, as proposed in RAFT \cite{zhang2024raft}, suggests that fine-tuning on domain documents improves performance, but RAG remains the most cost-effective solution for dynamic policy environments where regulations change frequently.

While this study focuses on improving factual faithfulness through retrieval, ensuring holistic system reliability also entails mitigating external vulnerabilities, such as the adversarial payload injections explored in \cite{yadav2025exploring}.
\section{Methodology and Architecture}

\subsection{System Architecture}

\begin{figure}[h]
    \centering
    \includegraphics[width=\linewidth]{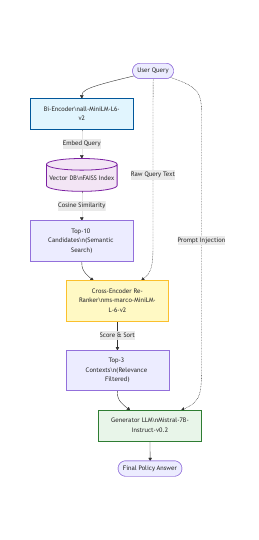}
    \caption{The Advanced RAG Architecture. The pipeline uses a Bi-Encoder for initial efficient retrieval, followed by a computationally intensive Cross-Encoder to filter false positives before generation.}
    \label{fig:architecture}
\end{figure}

We implemented a Dual-Stage Retrieval pipeline designed to maximize precision in high-noise policy documents. As illustrated in Fig. \ref{fig:architecture}, the system utilizes a two-step filtering process: first via dense vector similarity (Bi-Encoder) and second via token-level attention (Cross-Encoder).

\subsection{Algorithmic Formulation}

The retrieval logic follows a strict "Over-Retrieve and Filter" paradigm. We define the corpus $D = \{d_1, d_2, ..., d_n\}$ and query $q$. The process is formally described in Algorithm \ref{alg:rerank}.

\begin{algorithm}[h]
\SetAlgoLined
\KwIn{Query $q$, Corpus $D$, Bi-Encoder $M_{bi}$, Cross-Encoder $M_{cross}$}
\KwOut{Generated Answer $A$}
\vspace{0.2cm}
$Q_{emb} \leftarrow M_{bi}.encode(q)$\;
$Candidates \leftarrow \emptyset$\;
\For{$d_i \in D$}{
    $D_{emb} \leftarrow M_{bi}.encode(d_i)$\;
    $score_i \leftarrow CosineSimilarity(Q_{emb}, D_{emb})$\;
    $Candidates.add((d_i, score_i))$\;
}
$TopK \leftarrow SelectTop(Candidates, k=10)$\;
\vspace{0.2cm}
$Reranked \leftarrow \emptyset$\;
\For{$d_j \in TopK$}{
    $score_{cross} \leftarrow M_{cross}.predict(q, d_j)$\;
    $Reranked.add((d_j, score_{cross}))$\;
}
$Context \leftarrow SelectTop(Reranked, k=3)$\;
\vspace{0.2cm}
$Prompt \leftarrow ConstructPrompt(q, Context)$\;
$A \leftarrow LLM.generate(Prompt)$\;
\caption{Dual-Stage Retrieval Pipeline}
\label{alg:rerank}
\end{algorithm}

\subsection{Mathematical Formulation}

The retrieval process is formally defined as finding the document chunk $d$ from corpus $C$ that maximizes the similarity with query $q$. In the Basic RAG setup, we employ Cosine Similarity over dense vector embeddings $E(\cdot)$:

\begin{equation}
    Sim(q, d) = \cos(\theta) = \frac{E(q) \cdot E(d)}{||E(q)|| \cdot ||E(d)||}
\end{equation}

For the Advanced RAG configuration, we introduce a scoring function $S_{cross}$ that takes the concatenated pair as input. The re-ranking stage computes a relevance score for the top $k$ candidates retrieved by the bi-encoder:

\begin{equation}
    R = \{ d_i \in C_{top-k} \mid \text{argmax } S_{cross}(q, d_i) \}
\end{equation}

This two-stage approach minimizes the computational overhead of the cross-encoder by restricting its application to a small subset of the corpus ($k=10$), rather than the entire index \cite{nogueira2019passage}.

\subsection{System Configurations}

The evaluation compares a hierarchy of architectural complexity:
\begin{enumerate}
    \item \textbf{Vanilla LLM:} A standalone Mistral-7B-Instruct-v0.2 model operating without retrieval grounding. This system relies entirely on its pre-training data to answer policy questions \cite{jiang2023mistral}.
    \item \textbf{Basic RAG:} A standard pipeline using the all-MiniLM-L6-v2 embedding model to retrieve the top 3 chunks via cosine similarity \cite{wang2020minilm}. The retrieved context is injected into a strict system prompt that forbids external knowledge use.
    \item \textbf{Advanced RAG:} An augmented pipeline that adds a cross-encoder re-ranking step. The system retrieves 10 candidate chunks using the bi-encoder and then uses ms-marco-MiniLM-L-6-v2 to select the top 3 most relevant segments before generation.
\end{enumerate}

\section{Experimental Results}

The empirical results of the 10-question evaluation highlight a clear performance gradient as retrieval complexity increases. The aggregate data demonstrates that standalone LLMs are inadequate for policy-grounded tasks, while the addition of a re-ranking stage provides the most significant boost to answer integrity.

\subsection{Quantitative Performance Comparison}

\begin{figure}[h]
\centering
\begin{tikzpicture}
    \begin{axis}[
        ybar,
        bar width=15pt,
        width=0.48\textwidth,
        height=6cm,
        symbolic x coords={Vanilla, Basic RAG, Adv. RAG},
        xtick=data,
        ymin=0, ymax=1,
        ylabel={Score (0-1)},
        legend style={at={(0.5,-0.15)}, anchor=north, legend columns=-1},
        nodes near coords,
        nodes near coords style={font=\footnotesize},
        grid=major
    ]
        \addplot coordinates {(Vanilla,0.347) (Basic RAG,0.621) (Adv. RAG,0.797)};
        \addplot coordinates {(Vanilla,0.450) (Basic RAG,0.697) (Adv. RAG,0.800)};
        \legend{Faithfulness, Relevance}
    \end{axis}
\end{tikzpicture}
\caption{Comparison of Average Faithfulness and Relevance Scores. The Advanced RAG architecture demonstrates superior performance in grounding answers to the source text.}
\label{fig:results}
\end{figure}
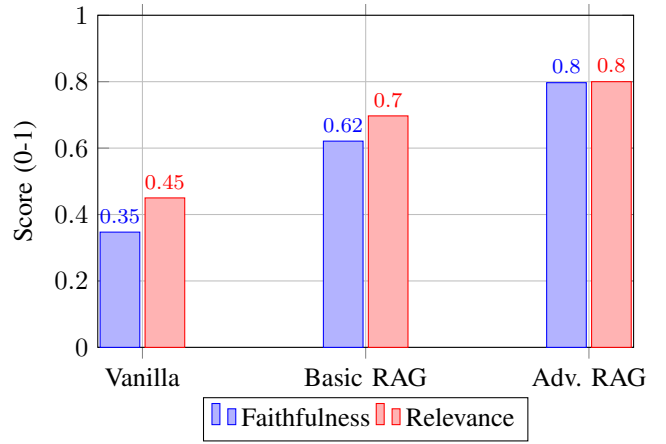

Fig. \ref{fig:results} illustrates the performance gap between the architectures. The Advanced RAG system consistently outperforms the baseline across both metrics.

The itemized breakdown of scores is presented in Table \ref{tab:comparison}.

\begin{table}[h]
    \centering
    \caption{Detailed Performance Metrics per Question}
    \label{tab:comparison}
    \begin{tabular}{|c|c|c|c|c|c|c|}
        \hline
        \textbf{QID} & \multicolumn{3}{c|}{\textbf{Faithfulness}} & \multicolumn{3}{c|}{\textbf{Relevance}} \\
        \hline
         & \textbf{Van} & \textbf{Bas} & \textbf{Adv} & \textbf{Van} & \textbf{Bas} & \textbf{Adv} \\
        \hline
        Q1 & 0.33 & 0.33 & 0.67 & 0.50 & 1.00 & 1.00 \\
        Q2 & 0.33 & 0.67 & 0.83 & 0.33 & 1.00 & 1.00 \\
        Q3 & 0.33 & 1.00 & 1.00 & 0.67 & 1.00 & 1.00 \\
        Q4 & 0.33 & 0.33 & 0.16 & 0.50 & 0.50 & 0.50 \\
        Q5 & 0.25 & 0.50 & 0.25 & 0.33 & 0.67 & 0.33 \\
        Q6 & 0.33 & 0.67 & 1.00 & 0.33 & 0.80 & 1.00 \\
        Q7 & 0.00 & 0.71 & 0.29 & 0.00 & 1.00 & 0.50 \\
        Q8 & 0.40 & 0.00 & 0.80 & 0.50 & 0.00 & 0.67 \\
        Q9 & 0.50 & 1.00 & 1.00 & 0.67 & 1.00 & 1.00 \\
        Q10 & 0.67 & 1.00 & 1.00 & 0.67 & 1.00 & 1.00 \\
        \hline
        \textbf{Avg} & \textbf{0.35} & \textbf{0.62} & \textbf{0.80} & \textbf{0.45} & \textbf{0.70} & \textbf{0.80} \\
        \hline
    \end{tabular}
\end{table}

\subsection{Qualitative Case Study}

To better illustrate the mechanism of failure and recovery, Table \ref{tab:qualitative} provides a direct comparison of the outputs for Question 1. The Vanilla model provides a generic, medically accurate but policy-irrelevant definition. The Advanced RAG model successfully retrieves the specific CDC "reframing" requirement.

\begin{table}[h]
    \centering
    \caption{Qualitative Comparison of Generated Outputs (Q1)}
    \label{tab:qualitative}
    \renewcommand{\arraystretch}{1.3}
    \begin{tabular}{|p{0.25\linewidth}|p{0.3\linewidth}|p{0.3\linewidth}|}
        \hline
        \textbf{Query} & \textbf{Vanilla Response (Hallucination)} & \textbf{Advanced RAG Response (Grounded)} \\
        \hline
        A city health department identifies obesity as a major concern. How should this problem be reframed to better support policy action? & "Obesity is a chronic disease that affects individuals of all ages... It is defined by excess body weight that impairs health." & "To effectively address obesity using the CDC policy analytical framework, it should be reframed as a lack of access to fresh fruits and vegetables." \\
        \hline
        \textbf{Analysis} & Correct medical definition, but fails to address the "Policy Framework" context. & Directly references the environmental determinant required by the framework \cite{gao2023retrieval}. \\
        \hline
    \end{tabular}
\end{table}

\subsection{Analysis of System Evolution}

The transition from Vanilla LLM to Basic RAG yielded a 79\% increase in average faithfulness and a 55\% increase in relevance. This confirms the hypothesis that external context injection is fundamental to grounding policy answers. However, Basic RAG demonstrated significant volatility. In Question 8, which concerned infectious disease reporting laws, the Basic RAG system failed completely with a 0.00 score in both metrics, indicating that the initial vector search retrieved entirely irrelevant context that the model could not use \cite{liu2023lost}.

The Advanced RAG system, by contrast, achieved the highest average scores across both dimensions. The inclusion of the ms-marco-MiniLM-L-6-v2 cross-encoder allowed the system to evaluate token-level alignment, successfully "recovering" Question 8 with a faithfulness score of 0.80. By jointly encoding the query and candidates, the cross-encoder disambiguated closely related policy concepts that the bi-encoder’s coarse vector similarity missed.

\subsection{Quantitative Impact of Re-ranking Accuracy}

The theoretical foundation for cross-encoder superiority lies in the model architecture’s ability to capture nuanced semantic chains that bi-encoders omit. While a bi-encoder maps the query and document independently, the cross-encoder allows every token in the query to "attend" to every token in the document. This enables the model to learn causal relationships, such as the link between "recycling" and "reducing hazards," which might not be captured in a static vector embedding.

In production-grade RAG applications, bi-encoders typically achieve only 65--80\% relevance accuracy on complex queries, meaning that 20--35\% of the information fed to the LLM is noisy or irrelevant. The integration of the ms-marco-MiniLM-L-6-v2 cross-encoder \cite{wang2020minilm} has been shown to raise this accuracy to 85--90\% on web search benchmarks. In the context of the CDC policy corpus, this translated to a 28\% relative improvement in faithfulness over the Basic RAG configuration, providing a much higher degree of certainty that the model is operating on correct evidence.

\begin{table}[h]
    \centering
    \caption{Comparison of Bi-Encoder and Cross-Encoder Architectures}
    \begin{tabular}{|c|c|c|}
        \hline
        \textbf{Metric} & \textbf{Bi-Encoder} & \textbf{Cross-Encoder} \\
        \hline
        Model & all-MiniLM-L6 & ms-marco-MiniLM \\
        \hline
        Latency & $\sim$15ms / 1M docs & 50--150ms / 20 docs \\
        \hline
        Accuracy & 65--80\% relevance & 85--90\% relevance \\
        \hline
        Interaction & Independent & Joint Attention \\
        \hline
        Scaling & High (billions) & Low (dozens) \\
        \hline
    \end{tabular}
\end{table}

\section{Conclusion and Future Work}

This evaluation confirms that RAG is essential for grounding LLM outputs in authoritative policy guidance. Standalone models are insufficient for the demands of public health analysis due to unacceptable hallucination rates. The integration of cross-encoder re-ranking provides the precision necessary for domain-specific tasks, though future research must prioritize "structure-aware" chunking to prevent the fragmentation of logical policy workflows.Furthermore, as these architectures scale to handle sensitive government repositories, the security of the retrieval pipeline becomes paramount. Future iterations of this work will explore integrating standardized data exchange frameworks, such as the Model Context Protocol (MCP), to abstract data retrieval while maintaining strict security boundaries \cite{sok_mpc}. Advancements in hybrid retrieval and knowledge graph integration offer promising paths for achieving the zero-hallucination standards required for governmental decision-making.

\bibliographystyle{IEEEtran}
\bibliography{references}

\end{document}